\pdfoutput=1
\documentclass{article} % For LaTeX2e
\usepackage{iclr2016_conference,times}

\usepackage{amssymb}
\usepackage{amsmath}
\usepackage{graphicx}
\usepackage{color}
\usepackage{subfig}

\title{Deep Linear Discriminant Analysis}

\author{Matthias Dorfer, Rainer Kelz \& Gerhard Widmer \thanks{\texttt{http://www.cp.jku.at/}} \\
Department of Computational Perception\\
Johannes Kepler University Linz\\
Linz, 4040, AUT \\
\texttt{\{matthias.dorfer, rainer.kelz, gerhard.widmer\}@jku.at} \\
}

\iclrfinalcopy % Uncomment for camera-ready version

\begin{document}

\maketitle

\begin{abstract}
%!TEX root = iclr2016_conference.tex

We introduce Deep Linear Discriminant Analysis (\emph{DeepLDA}) which learns linearly separable latent representations in an end-to-end fashion.
Classic LDA extracts features which preserve class separability and is used for dimensionality reduction for many classification problems. 
The central idea of this paper is to put LDA on top of a deep neural network.
This can be seen as a non-linear extension of classic LDA.
Instead of maximizing the likelihood of target labels for individual samples,
we propose an objective function that pushes the network to produce feature distributions which:
(a) have low variance within the same class and (b) high variance between different classes.
Our objective is derived from the general LDA eigenvalue problem
and still allows to train with stochastic gradient descent and back-propagation.
For evaluation we test our approach on three different benchmark datasets (MNIST, CIFAR-10 and STL-10).
DeepLDA produces competitive results on MNIST and CIFAR-10
and outperforms a network trained with categorical cross entropy (having the same architecture) on a supervised setting of STL-10.

\end{abstract}

\section{Introduction}
\label{sec:introduction}
%!TEX root = iclr2016_conference.tex

Linear Discriminant Analysis (LDA) is a method from multivariate statistics
which seeks to find a linear projection of high-dimensional observations into a lower-dimensional space \citep{Fisher1936LDA}.
When its preconditions are fulfilled, LDA allows to define optimal linear decision boundaries in the resulting latent space.
The aim of this paper is to exploit the beneficial properties of classic LDA
(low intra class variability, hight inter-class variability, optimal decision boundaries)
by reformulating its objective to learn linearly separable representations based on a deep neural network (DNN).

Recently, methods related to LDA achieved great success in combination with deep neural networks.
Andrew et al. published a deep version of Canonical Correlation Analysis (DCCA) \citep{Andrew2013DCCA}.
In their evaluations, DCCA is used to produce correlated representations of multi-modal input data of simultaneously recorded acoustic and articulatory speech data.
Clevert et al. propose Rectified Factor Networks (RFNs) which are a neural network interpretation of classic factor analysis \citep{Clever2015RFNs}. RFNs are used for unsupervised pre-training and help to improve classification performance on four different benchmark datasets.
A similar method called PCANet -- as well as an LDA based variation -- was proposed by \cite{Chan_2015_PCANet}.
PCANet can be seen as a simple unsupervised convolutional deep learning approach.
The method proceeds with cascaded Principal Component Analysis (PCA), binary hashing and block histogram computations.
However, one crucial bottleneck of their approach is its limitation to very shallow architectures (two stages) \citep{Chan_2015_PCANet}.

Stuhlsatz et. al. already picked up the idea of combining LDA with a neural networks and proposed a generalized version of LDA \citep{Stuhlsatz2012LDA}.
Their approach starts with pre-training a stack of restricted Boltzmann machines.
In a second step, the pre-trained model is fine-tuned with respect to a linear discriminant criterion.
LDA has the disadvantage that it overemphasises large distances at the cost of confusing neighbouring classes.
In \citep{Stuhlsatz2012LDA} this problem is tackled by a heuristic weighting scheme for computing the within-class scatter matrix required for LDA optimization.

\subsection{Main Idea of this Paper}
The approaches mentioned so far all have in common
that they are based on well established methods from multivariate statistics.
Inspired by their work, we propose an end-to-end DNN version of LDA - namely \textit{Deep Linear Discriminant Analysis} (\textit{DeepLDA}).

Deep learning has become the state of the art in automatic feature learning
and replaced existing approaches based on hand engineered features in many fields such as object recognition \citep{krizhevsky2012imagenet}.
DeepLDA is motivated by the fact that when the preconditions of LDA are met, it is capable of finding linear combinations of the input features which allow for optimal linear decision boundaries.
In general, LDA takes features as input.
The intuition of our method is to use LDA as an objective on top of a powerful feature learning algorithm.
Instead of maximizing the likelihood of target labels for individual samples,
we propose an LDA eigenvalue-based objective function that pushes the network to produce discriminative feature distributions.
The parameters are optimized by back-propagating the error of an LDA-based objective through the entire network.
We tackle the feature learning problem by focusing on directions
in the latent space with smallest discriminative power.
This replaces the weighting scheme of \citep{Stuhlsatz2012LDA}
and allows to operate on the original formulation of LDA.
We expect that DeepLDA will produce linearly separable hidden representations with similar discriminative power in all directions of the latent space. Such representations should also be related with a high classification potential of the respective networks.
The experimental classification results reported below will confirm this
positive effect on classification accuracy, and two additional experiments
(Section \ref{sec:investigations}) will give us some first qualitative confirmation
that the learned representations show the expected properties.

The reminder of the paper is structured as follows.
In Section \ref{sec:dnn} we provide a general formulation of a DNN.
Based on this formulation we introduce DeepLDA,
a non-linear extension to classic LDA in Section \ref{sec:dlda}.
In Section \ref{sec:experiments} we experimentally evaluate our approach on three benchmark datasets.
Section \ref{sec:investigations} provides a deeper insight into the structure of DeepLDA's internal represenations.
In Section \ref{sec:conclusion} we conclude the paper.

\section{Deep Neural Networks}
\label{sec:dnn}
%!TEX root = iclr2016_conference.tex

As the proposed model is built on top of a DNN
we briefly describe the training paradigm of a network used for classification problems such as object recognition.

\begin{figure*}[ht]
\label{fig:comparison_lda_cce}
\centering
\subfloat[The output of the network gets normalized by a soft max layer to form valid probabilities. The CCE objective maximizes the likelihood of the target class under the model.]{\label{fig:cce_sketch}{\includegraphics[width=0.44\textwidth]{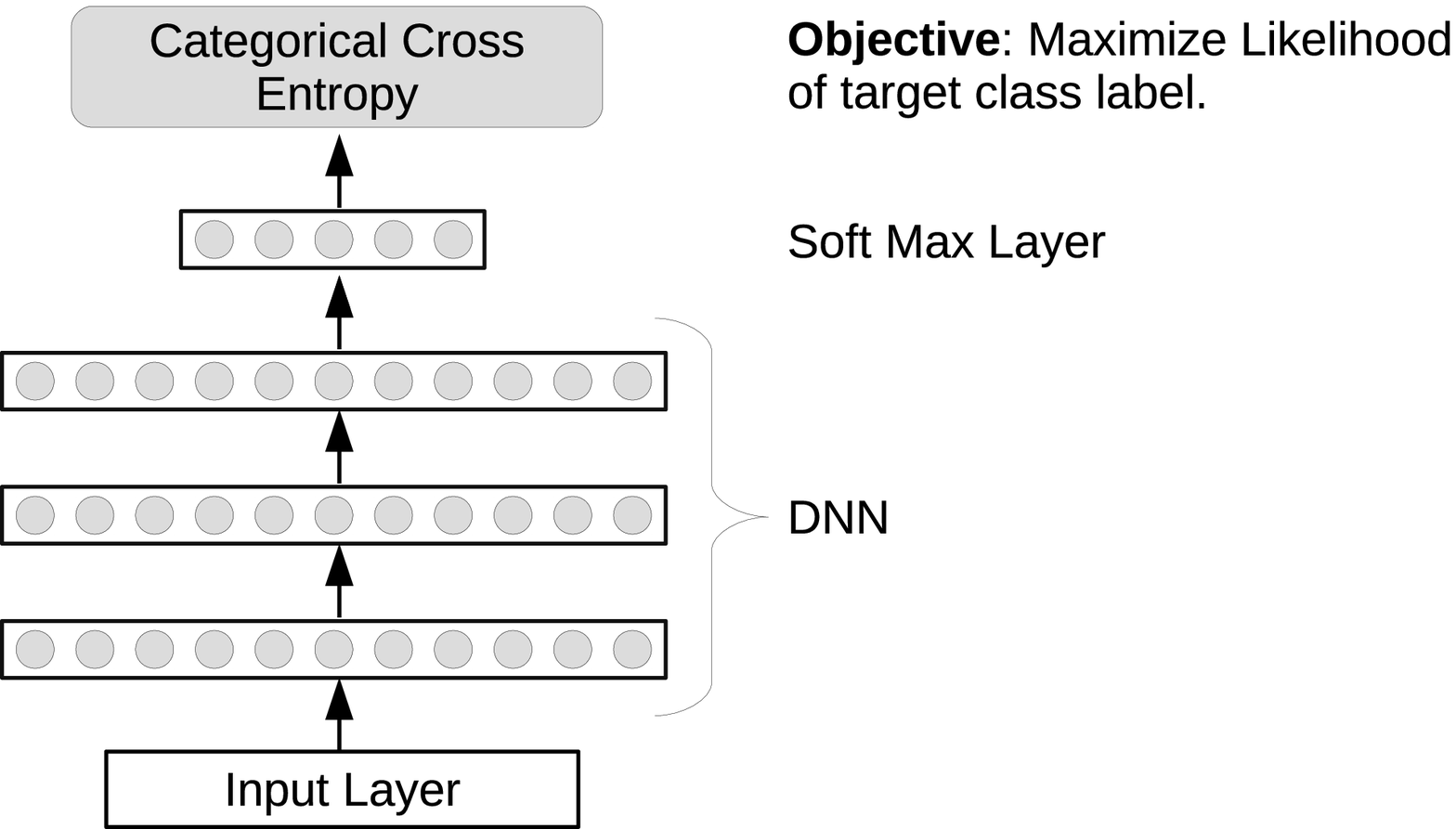} }}%
\qquad
\subfloat[On the topmost hidden layer we compute an LDA which produces corresponding eigenvalues.
The optimization target is to maximize those eigenvalues.]{\label{fig:lda_sketch}{\includegraphics[width=0.44\textwidth]{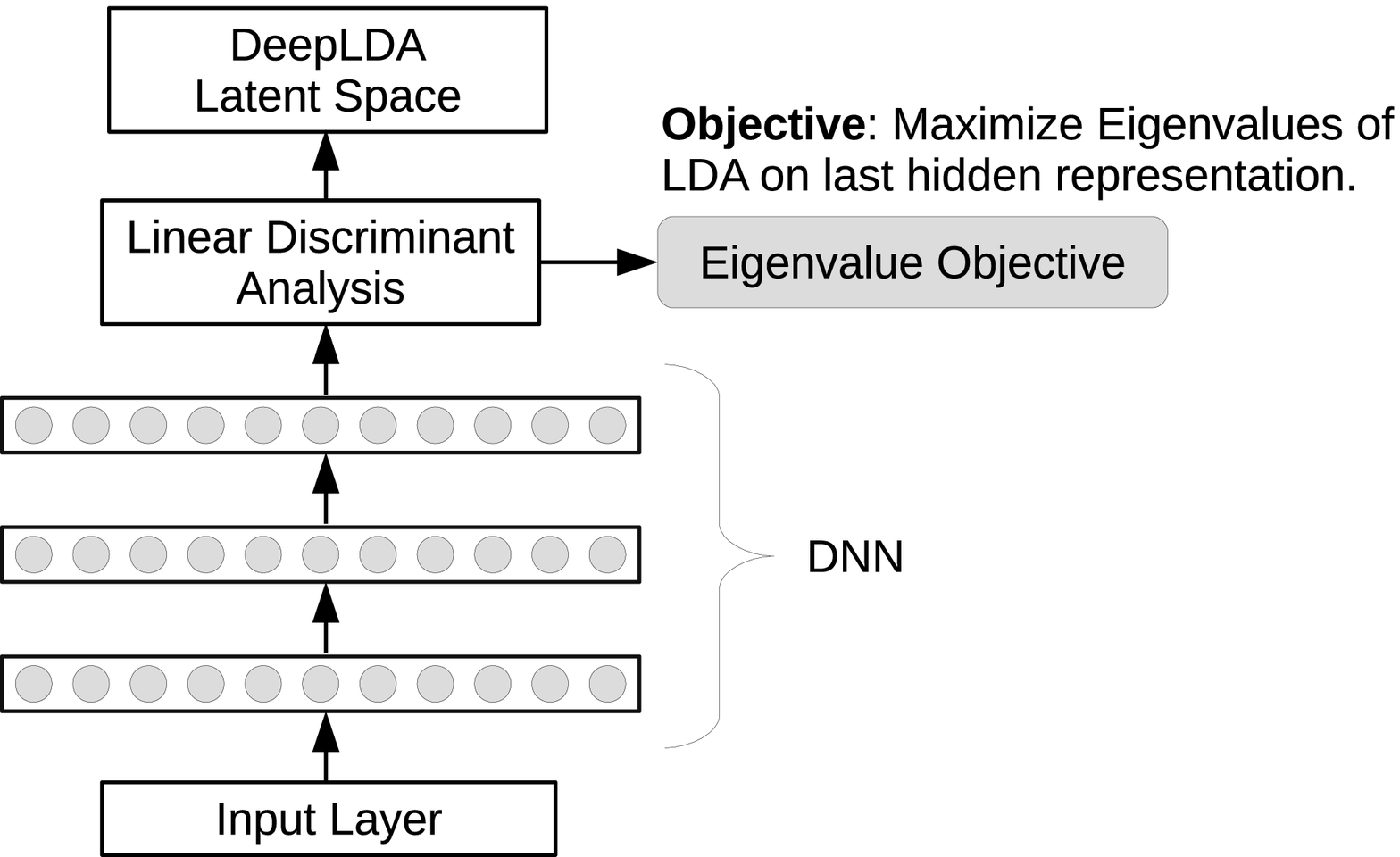} }}%
\caption{Schematic sketch of a DNN and DeepLDA. For both architectures the input data is first propagated through the layers of the DNN. However, the final layer and the optimization target are different.}
\end{figure*}

A neural network with $P$ hidden layers is represented as a non-linear function $f(\Theta)$ with model parameters $\Theta=\{\Theta_1,...,\Theta_P\}$.
In the supervised setting we are additionally given a set of $N$ train samples $\mathbf{x}_1,...\mathbf{x}_N$ along with corresponding classification targets $t_1,...t_N \in \{1,...,C\}$.
We further assume that the network output $\mathbf{p}_i=(p_{i,1},...,p_{i,C})=f(\mathbf{x}_i,\Theta)$ is normalized by the softmax-function to obtain class (pseudo-)probabilities.
The network is then optimized using Stochastic Gradient Descent (SGD) with the goal of finding an optimal model parametrization $\Theta$ with respect to a certain loss function $l_i(\Theta) = l(f(\mathbf{x}_i,\Theta), t_i)$.
\begin{equation}
\Theta = \underset{\Theta}{\arg \, \min} \frac{1}{N} \sum_{i=1}^{N} l_i(\Theta)
\end{equation}
For multi-class classification problems, Categorical-Cross-Entropy (CCE) is a commonly used optimization target and formulated for observation $\mathbf{x}_i$ and target label $t_i$ as follows
\begin{equation}
l_i(\Theta) = - \sum_{j=1}^{C} y_{i,j} log(p_{i,j})
\end{equation}
where $y_{i,j}$ is $1$ if observation $\mathbf{x}_i$ belongs to class $t_i$ ($j=t_i$) and $0$ otherwise.
In particular, the CCE tries to maximize the likelihood of the target class $t_i$ for each of the individual training examples $\mathbf{x}_i$ under the model with parameters $\Theta$.
Figure \ref{fig:cce_sketch} shows a sketch of this general network architecture.

We would like to emphasize that objectives such as CCE do not impose any direct constraints -- such as linear separability -- on the latent space representation.

\section{Deep Linear Discriminant Analysis (DeepLDA)}
\label{sec:dlda}
In this section we first provide a general introduction to LDA.
Based on this introduction we propose DeepLDA, which optimizes an LDA-based optimization target in an end-to-end DNN fashion.
Finally we describe how DeepLDA is used to predict class probabilities of unseen test samples.

\subsection{Linear Discriminant Analysis}
\label{subsec:lda_general}
%!TEX root = iclr2016_conference.tex

Let ${\mathbf{x}_1, ..., \mathbf{x}_N} = \mathbf{X} \in \mathbb{R}^{N \times d}$ denote a set of $N$ samples belonging to $C$ different classes $c \in \{1, ..., C\}$.
The input representation $\mathbf{X}$ can either be hand engineered features,
or hidden space representations $\mathbf{H}$ produced by a DNN \citep{Andrew2013DCCA}.
LDA seeks to find a linear projection $\mathbf{A} \in \mathbb{R}^{l \times d}$ into a lower $l$-dimensional subspace $L$ where $l=C-1$.
The resulting linear combinations of features $\mathbf{x}_i \mathbf{A}^T$ are maximally separated in this space \citep{Fisher1936LDA}.
The LDA objective to find projection matrix $\mathbf{A}$ is formulated as:
\begin{equation}
\label{eq:argmax}
\underset{\mathbf{A}}{\arg\max} \frac{|\mathbf{A} S_b \mathbf{A}^T|}{|\mathbf{A} S_w \mathbf{A}^T|}
\end{equation}

where $\mathbf{S}_b$ is the between scatter matrix and defined via the total scatter matrix $\mathbf{S}_t$ and within scatter matrix $\mathbf{S}_w$ as $\mathbf{S}_b=\mathbf{S}_t - \mathbf{S}_w$.
$\mathbf{S}_w$ is defined as the mean of the $C$ individual class covariance matrices $\mathbf{S}_c$ (Equation (\ref{eq:Sc}) and (\ref{eq:Sw})).
$\mathbf{\bar{X}}_c = \mathbf{X}_c - \mathbf{m}_c$ are the mean-centered observations of class $c$ with per-class mean vector $\mathbf{m}_c$ ($\mathbf{\bar{X}}$ is defined analogously for the entire population $\mathbf{X}$).
The total scatter matrix $\mathbf{S}_t$ is the covariance matrix over the entire population of observations $\mathbf{X}$.
\begin{equation}
\label{eq:Sc}
\mathbf{S}_c = \frac{1}{N_c-1} \mathbf{\bar{X}}_c^T \mathbf{\bar{X}}_c
\end{equation}
\begin{equation}
\label{eq:Sw}
\mathbf{S}_w = \frac{1}{C} \sum_{c} \mathbf{S}_c
\end{equation}
\begin{equation}
\label{eq:St}
\mathbf{S}_t = \frac{1}{N-1} \mathbf{\bar{X}}^T \mathbf{\bar{X}}
\end{equation}

The linear combinations that maximize the objective in Equation (\ref{eq:argmax}) maximize the ratio of between- and within-class scatter also reffered to as separation.
This means in particular that a set of projected observations of the same class show low variance, whereas the projections of observations of different classes have high variance in the resulting space $L$.
To find the optimum solution for Equation (\ref{eq:argmax}) one has to solve the general eigenvalue problem $\mathbf{S}_b \mathbf{e} = \mathbf{v} \mathbf{S}_w \mathbf{e}$.
The projection matrix $\mathbf{A}$ is the set of eigenvectors $\mathbf{e}$ associated with this problem.
In the following sections we will cast LDA as an objective function for DNN.

\subsection{DeepLDA Model Configuration}
\label{subsec:dlda_scetch}
%!TEX root = iclr2016_conference.tex

Figure \ref{fig:lda_sketch} shows a schematic sketch of DeepLDA.
Instead of sample-wise optimization of the CCE loss on the predicted class probabilities
(see Section \ref{sec:dnn}) we put an \emph{LDA-layer} on top of the DNN.
This means in particular that we do not penalize the misclassification of individual samples.
Instead we try to produce features that show a low intra-class and high inter-class variability.
We address this maximization problem by a modified version of the general LDA eigenvalue problem proposed in the following section.
In contrast to CCE, DeepLDA optimization operates on the properties of the distribution parameters of the hidden representation produced by the neural net.
As eigenvalue optimization is tied to its corresponding eigenvectors (a linear projection matrix), DeepLDA can be also seen as a special case of a dense layer.

%This means we seek a parametrization $\Theta$ of the net
%which allows a projection of the last internal (hidden) representation
%towards an optimally linearly separable space.
%All required non-linear transformations have to be therefore applied to the data
%when travelling through the layers of the neural net.

\subsection{Modified DeepLDA Optimization Target}
\label{subsec:lda_modified}
%!TEX root = iclr2016_conference.tex

Based on Section \ref{subsec:lda_general} we reformulate the LDA objective to be suitable for a combination with deep learning.
As already discussed by \cite{Stuhlsatz2012LDA} and \cite{Lu2005LDAregularization} the estimation of $\mathbf{S}_w$ overemphasises high eigenvalues whereas small eigenvalues are estimated as too low.
To weaken this effect, \cite{Friedman1989regularizedLDA} proposed to regularize the within scatter matrix by adding a multiple of the identity matrix $\mathbf{S}_w + \lambda \mathbf{I}$.
Adding the identity matrix has the second advantage of stabilizing small eigenvalues.
The resulting eigenvalue problem is then formulated as
\begin{equation}
\label{eq:eigval_problem2}
\mathbf{S}_b \mathbf{e}_i = v_i (\mathbf{S}_w + \lambda \mathbf{I}) \mathbf{e }_i
\end{equation}
where $\mathbf{e}=\mathbf{e}_1,...,\mathbf{e}_{C-1}$ are the resulting eigenvectors and $\mathbf{v}=v_1,...v_{C-1}$ the corresponding eigenvalues.
Once the problem is solved, each eigenvalue $v_i$ quantifies the amount of discriminative variance (separation) in direction of the corresponding eigenvector $\mathbf{e}_i$.
If one would like to combine this objective with a DNN the optimization target
would be the maximization of the individual eigenvalues.
In particular, we expect that maximizing the individual eigenvalues -- which reflect the separation in the respective eigenvector directions -- leads to a maximization of the discriminative power of the neural net.
In our initial experiments we started to formulate the objective as:
\begin{equation}
\label{eq:lda_obj_mean}
\underset{\Theta}{\arg \max} \frac{1}{C-1} \sum_{i=1}^{C-1}v_i
\end{equation}
One problem we discovered with the objective in Equation (\ref{eq:lda_obj_mean}) is that the net favours trivial solutions e.\ g.\ maximize only the largest eigenvalue as this produces the highest reward.
In terms of classification this means that it maximizes the distance of classes that are already separated at the expense of -- potentially non-separated -- neighbouring classes.
This was already discussed by \citep{Stuhlsatz2012LDA}
and tackled by a weighted computation of the between scatter matrix $\mathbf{S}_b$.

We propose a different solution to this problem and address it by focusing
our optimization on the smallest of all $C-1$ available eigenvalues.
In particular we consider only the $k$ eigenvalues that do not exceed a certain threshold for variance maximization:
\begin{equation}
\label{eq:lda_loss}
\underset{\Theta}{\arg \max} \frac{1}{k} \sum_{i=1}^{k}v_i \; \; \textnormal{with} \; \; \{v_1,...,v_k\} = \{v_j | v_j < \min\{v_1,...,v_{C-1}\} + \epsilon\}
\end{equation}
The intuition behind this formulation is to learn a net parametrization that
pushes as much discriminative variance as possible into all of the $C-1$ available feature dimensions.

We would like to underline that this formulation allows to train DeepLDA networks with back-propagation in end-to-end fashion (see Appendix for a derivative of the loss functions's gradient).
Our models are optimized with the Nesterov momentum version of mini-batch SGD.
Related methods already showed that mini-batch learning on distribution parameters (in this case covariance matrices) is feasible
if the batch-size is sufficiently large to be representative for the entire population \citep{Wang_2015_MultiView, Wang2015unsupervised}.

\subsection{Classification by DeepLDA}
\label{subsec:dlda_clf}
%!TEX root = iclr2016_conference.tex

This section describes how the most likely class label is assigned to an unseen test sample $\mathbf{x}_t$ once the network is trained and parametrized.
In a first step we compute the topmost hidden representation $\mathbf{H}$ on the entire training set $\mathbf{X}$.
On this hidden representation we compute the LDA as described in Section \ref{subsec:lda_general} and \ref{subsec:lda_modified} producing the corresponding eigenvectors $\mathbf{e} = \{\mathbf{e}_i\}_{i=1}^{C-1}$ which form the LDA projection matrix $\mathbf{A}$.
We would like to emphasize that since the parameters of the network are fixed at this stage we make use of the entire training set to provide a stable estimate of the LDA projection.
Based on $\mathbf{A}$ and the per-class mean hidden representations $\bar{\mathbf{H}}_c=(\bar{\mathbf{h}}_1^T,...,\bar{\mathbf{h}}_C^T)$ the distances of sample $\mathbf{h}_t$ to the linear decision hyperplanes \citep{Friedman2001Elements} are defined as
\begin{equation}
\mathbf{d} = \mathbf{h}^T_t \mathbf{T}^T
- \frac{1}{2} diag \left( \bar{\mathbf{H}}_c \mathbf{T}^T \right)
\text{with } \mathbf{T}= \bar{\mathbf{H}}_c\mathbf{A}\mathbf{A}^T
\end{equation}
where $\mathbf{T}$ are the decision hyperplane normal vectors.
The subtracted term is the bias of the decision functions placing the decision boundaries in between the means of the respective class hidden representations (no class priors included).
The vector of class probabilities for test sample $\mathbf{x}_t$ is then computed by applying the logistic function $\mathbf{p}_c' = 1 / (1 + \mathrm{e}^{-\mathbf{d}})$
and further normalized by $\mathbf{p}_c = \mathbf{p}_c'/\sum p_i'$ to sum to one.
Finally we assign class $i$ with highest probability as $\arg \max_i p_i$ to the unseen test sample $\mathbf{x}_t$.

% The model fits a Gaussian density to each class,
% assuming that all classes share the same covariance matrix.

\section{Experiments}
\label{sec:experiments}
%!TEX root = iclr2016_conference.tex

\begin{figure*}[ht]
\centering
\subfloat[ ]{{\includegraphics[width=0.053\textwidth]{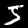} }}%
\subfloat[ ]{{\includegraphics[width=0.053\textwidth]{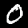} }}%
\qquad
\subfloat[ ]{{\includegraphics[width=0.06\textwidth]{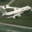} }}%
\subfloat[ ]{{\includegraphics[width=0.06\textwidth]{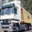} }}%
\qquad
\subfloat[ ]{{\includegraphics[width=0.18\textwidth]{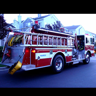} }}%
\subfloat[ ]{{\includegraphics[width=0.18\textwidth]{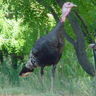} }}%
\caption{Example images of evaluation data sets (a)(b) MNIST, (c)(d) CIFAR-10, (e)(f) STL-10.
The relative size differences between images from the three data sets are kept in this visualization.}
\label{fig:datasets}
\end{figure*}

In this section we present an experimental evaluation of DeepLDA on three benchmark data sets -- namely MNIST, CIFAR-10 and STL-10 (see Figure \ref{fig:datasets} for some sample images).
We compare the results of DeepLDA with the CCE based optimization target as well as the present state of the art of the respective datasets.
In addition, we provide details on the network architectures, hyper parameters and respective training/optimization approaches used in our experiments.

\subsection{Experimental Setup}
%!TEX root = iclr2016_conference.tex

The general structure of the networks is similar for all of the three datasets
and identical for CIFAR-10 and STL-10.
The architecture follows the VGG model with sequences of $3\times3$ convolutions \citep{simonyan2014very}.
Instead of a dense classification layer we use global average pooling on the feature maps of the last convolution layer \citep{LinCY2013NIN}.
We picked this architecture as it leads to well-posed problems for covariance estimation: many samples vs. low feature space dimension.
We further apply batch normalization \citep{LoffeS2015BatchNorm} after each convolutional layer which (1) helped to increase convergence speed and (2) improved the performance of all our models. Batch normalization has a positive effect on both CCE as well as DeepLDA-based optimization.
In Table \ref{tab:model_architecture} we outline the structure of our models in detail.
All networks are trained using SGD with Nesterov momentum.
The initial learning rate is set to $0.1$ and the momentum is fixed at $0.9$ for all our models.
The learning rate is then halved every 25 epochs for CIFAR-10 and STL-10 and every 10 epochs for MNIST.
For further regularization we add weight decay with a weighting of $0.0001$ on all trainable parameters of the models.
The between-class covariance matrix regularization weight $\lambda$ (see Section \ref{subsec:lda_modified}) is set to $0.001$ and the $\epsilon$-offset for DeepLDA to $1$.

One hyper-parameter that varies between the datasets is the batch size used for training DeepLDA.
Although a large batch size is desired to get stable covariance estimates
it is limited by the amount of memory available on the GPU.
The mini-batches for DeepLDA were for MNIST: 1000, for CIFAR-10: 1000 and for STL-10: 200.
For CCE training, a batch size of 128 is used for all datasets.
The models are trained on an NVIDIA Tesla K40 with 12GB of GPU memory.

\begin{table}[t]
\caption{Model Specifications. BN: Batch Normalization, ReLu: Rectified Linear Activation Function, CCE: Categorical Cross Entropy. The mini-batch sizes of DeepLDA are: MNIST(1000), CIFAR-10(1000), STL-10(200). For CCE training a constant batch size of 128 is used.}
\label{tab:model_architecture}
\begin{center}
\begin{tabular}{c|c}
\hline
\multicolumn{1}{c}{CIFAR-10 and STL-10}  & \multicolumn{1}{c}{MNIST} \\
\hline \\
Input $3\times32\times32$ ($96\times96$)	& Input $1\times28\times28$ \\
\hline
\multicolumn{2}{c}{$3\times3$ Conv(pad-1)-$64$-BN-ReLu} \\
\multicolumn{2}{c}{$3\times3$ Conv(pad-1)-$64$-BN-ReLu} \\
\multicolumn{2}{c}{$2\times2$ Max-Pooling + Drop-Out($0.25$)} \\
\hline
$3\times3$ Conv(pad-1)-$128$-BN-ReLu 	& $3\times3$ Conv(pad-1)-$96$-BN-ReLu \\
$3\times3$ Conv(pad-1)-$128$-BN-ReLu		& $3\times3$ Conv(pad-1)-$96$-BN-ReLu \\
$2\times2$ Max-Pooling + Drop-Out($0.25$)		& $2\times2$ Max-Pooling + Drop-Out($0.25$) \\
\hline
$3\times3$ Conv(pad-1)-$256$-BN-ReLu \\
$3\times3$ Conv(pad-1)-$256$-BN-ReLu \\
$3\times3$ Conv(pad-1)-$256$-BN-ReLu \\
$3\times3$ Conv(pad-1)-$256$-BN-ReLu \\
$2\times2$ Max-Pooling + Drop-Out($0.25$) \\
\hline
$3\times3$ Conv(pad-0)-$1024$-BN-ReLu 	& $3\times3$ Conv(pad-0)-$256$-BN-ReLu \\
Drop-Out($0.5$)								& Drop-Out($0.5$) \\
\hline
$1\times1$ Conv(pad-0)-$1024$-BN-ReLu	& $1\times1$ Conv(pad-0)-$256$-BN-ReLu \\
Drop-Out($0.5$) 								& Drop-Out($0.5$)\\
\hline
$1\times1$ Conv(pad-0)-$10$-BN-ReLu & $1\times1$ Conv(pad-0)-$10$-BN-ReLu \\
$2\times2$ ($10\times10$) Global-Average-Pooling & $5\times5$ Global-Average-Pooling\\
\hline
\multicolumn{2}{c}{Soft-Max with CCE or LDA-Layer} \\
\end{tabular}
\end{center}
\end{table}

\subsection{Experimental Results}
%!TEX root = iclr2016_conference.tex

We describe the benchmark datasets as well as the pre-processing and data augmentation used for training.
We present our results and relate them to the present state of the art for the respective dataset.
As DeepLDA is supposed to produce a linearly separable feature space,
we also report the results of a linear Support Vector Machine trained on the latent space of DeepLDA (tagged with \emph{LinSVM}).
The results of our network architecture trained with CCE are marked as \emph{OurNetCCE}.
To provide a complete picture of our experimental evaluation we also show classification results
of an LDA on the topmost hidden representation of the networks trained with CCE (tagged with \emph{OurNetCCE(LDA)}).

\subsubsection{MNIST}
The MNIST dataset consists of $28\times28$ gray scale images of handwritten digits ranging from 0 to 9.
The dataset is structured into 50000 train samples, 10000 validation samples and 10000 test samples.
For training we did not apply any pre-processing nor data augmentation.
We present results for two different scenarios.
In scenario \textit{MNIST-50k} we train on the 50000 train samples
and use the validation set to pick the parametrization which produces the best results on the validation set.
In scenario \textit{MNIST-60k} we train the model for the same number of epochs as in \textit{MNIST-50k} but also use the validation set for training.
Finally we report the accuracy of the model on the test set after the last training epoch.
This approach was also applied in \citep{LinCY2013NIN} which produce state of the art results on the dataset.

Table \ref{tab:res_mnist} summarizes all results on the MNIST dataset.
DeepLDA produces competitive results -- having a test set error of $0.29\%$ -- although no data augmentation is used.
In the approach described in \citep{Graham2014SparseConv} the train set is extended with translations of up to two pixels.
We also observe that a linear SVM trained on the learned representation produces comparable results on the test set.
It is also interesting that early stopping with best-model-selection (MNIST-50k) performs better than training on MNIST-60k even though 10000 more training examples are available.

\begin{table}[t]
\caption{Comparison of test errors on MNIST}
\label{tab:res_mnist}
\begin{center}
\begin{tabular}{ll}
\hline
\multicolumn{1}{c}{Method}  & \multicolumn{1}{l}{Test Error}
\\ \hline
NIN + Dropout (\cite{LinCY2013NIN})		& $0.47\%$ \\
Maxout (\cite{goodfellow2013maxout}) 	& $0.45\%$ \\
DeepCNet(5,60) (\cite{Graham2014SparseConv})	& $0.31\%$ (train set translation) \\
\hline
OurNetCCE(LDA)-50k      	& $0.39\%$ \\
OurNetCCE-50k      		& $0.37\%$ \\
OurNetCCE-60k      		& $0.34\%$ \\
DeepLDA-60k      		& $0.32\%$ \\
OurNetCCE(LDA)-60k      	& $0.30\%$ \\
DeepLDA-50k      		& $\mathbf{0.29\%}$ \\
DeepLDA-50k(LinSVM)		& $\mathbf{0.29\%}$ \\
\hline
\end{tabular}
\end{center}
\end{table}

\subsubsection{CIFAR-10}
\label{subsec:cifar10_results}
The CIFAR-10 dataset consists of tiny $32\times32$ natural RGB images containing samples of 10 different classes.
The dataset is structured into 50000 train samples and 10000 test samples.
We pre-processed the dataset using global contrast normalization and ZCA whitening as proposed by \cite{goodfellow2013maxout}.
During training we only apply random left-right flips on the images -- no additional data augmentation is used.
In training, we follow the same procedure as described for the MNIST dataset above to make use of the entire 50000 train images.

Table \ref{tab:res_cifar10} summarizes our results and relates them to the present state of the art.
Both OurNetCCE and DeepLDA produce state of the art results on the dataset when no data augmentation is used.
Although DeepLDA performs slightly worse than CCE it is capable of producing competitive results on CIFAR-10.

\begin{table}[t]
\caption{Comparison of test errors on CIFAR-10}
\label{tab:res_cifar10}
\begin{center}
\begin{tabular}{ll}
\hline
\multicolumn{1}{c}{Method}  & \multicolumn{1}{l}{Test Error}
\\ \hline
NIN + Dropout (\cite{LinCY2013NIN}) 				& $10.41\%$ \\
Maxout (\cite{Graham2014SparseConv})      		& $9.38\%$ \\
NIN + Dropout (\cite{LinCY2013NIN}) 				& $8.81\%$ (data augmentation) \\
DeepCNINet(5,300) (\cite{Graham2014SparseConv}) 	& $\mathbf{6.28\%}$ (data augmentation) \\
\hline
DeepLDA(LinSVM)			& $7.58\%$ \\
DeepLDA	      			& $7.29\%$ \\
OurNetCCE(LDA)	      	& $7.19\%$ \\
OurNetCCE	      		& $7.10\%$ \\
\hline
\end{tabular}
\end{center}
\end{table}

\subsubsection{STL-10}
\label{subsec:stl10_results}
Like CIFAR-10, the STL-10 data set contains natural RGB images of 10 different object categories.
However, with $96\times96$ pixels the size of the images is larger,
and the training set is considerably smaller, with only 5000 images.
The test set consists of 8000 images.
In addition, STL-10 contains 100000 unlabelled images but we do not make use of this additional data at this point
as our approach is fully supervised.
For that reason we first perform an experiment (\emph{Method-4k}) where we do not follow the evaluation strategy described in \citep{Coates2011STL},
where models are trained on 1000 labeled and 100000 unlabeled images.
Instead, we directly compare CCE and DeepLDA in a fully supervised setting.
As with MNIST-50k we train our models on 4000 of the train images and use the rest (1000 images) as a validation set to pick the best performing parametrization.
The results on the \emph{Method-4k-Setting} of STL-10 are presented in the top part of Table \ref{tab:res_stl10}.
Our model trained with CCE achieves an accuracy of $78.39\%$.
The same architecture trained with DeepLDA improves the test set accuracy by more than 3 percentage points and achieves $81.46\%$.
In our second experiment (\emph{Method-1k}) we follow the evaluation strategy described in \citep{Coates2011STL} but without using the unlabelled data.
We train our models on the 10 pre-defined folds (each fold contains 1000 train images) and report the average accuracy on the test set.
The model optimized with CCE (\emph{OurNetCCE-1k}) achieves $57.44\%$ accuracy on the test set which is in line with the supervised results reported in \citep{zhao2015stackedwhatwhere}.

Our model trained with DeepLDA achieves $66.97\%$ average test set accuracy.
This is a performance gain of $9.53\%$ in contrast to CCE and
it shows that the advantage of DeepLDA compared to CCE becomes even more apparent when the amount of labeled data is low.
When comparing \emph{DeepLDA-1k} with LDA applied on the features computed by a network trained with CCE (\emph{OurNetCCE(LDA)-1k}, $59.48\%$),
we find that the end-to-end trained LDA-features outperform the standard CCE approach.
A direct comparison with state of the art results as reported in \citep{zhao2015stackedwhatwhere, swersky2013MultiTask, dosovitskiy2014discriminative} is not possible
because these models are trained under semi-supervised conditions using both unlabelled and labelled data.
However, the results suggest that a combination of DeepLDA with methods such as proposed by \cite{zhao2015stackedwhatwhere} is a very promising future direction.

\begin{table}[t]
\caption{Comparison of test set accuracy on a purely supervised setting of STL-10. (\emph{Method-4k}: 4000 train images, \emph{Method-1k}: 1000 train images.)}
\label{tab:res_stl10}
\begin{center}
\begin{tabular}{ll}
\hline
\multicolumn{1}{c}{Method-4k}  & \multicolumn{1}{l}{Test Accuracy-4k}
\\ \hline
OurNetCCE(LDA)-4k	& $78.50\%$ \\
OurNetCCE-4k			& $78.84\%$ \\
DeepLDA-4k			& $81.16\%$ \\
DeepLDA(LinSVM)-4k	& $\mathbf{81.40\%}$ \\
\\
\hline
\multicolumn{1}{c}{Method-1k}  & \multicolumn{1}{l}{Test Accuracy-1k}
\\ \hline
%Multi-Task Bayesian Optimization (\cite{swersky2013MultiTask})	& $70.10\%$ \\
%Exemplar Convnets (\cite{dosovitskiy2014discriminative})			& $75.40\%$ \\
SWWAE (\cite{zhao2015stackedwhatwhere}) 		& $57.45\%$ \\
SWWAE (\cite{zhao2015stackedwhatwhere}) 		& $74.33\%$ (semi-supervised) \\
\hline
DeepLDA(LinSVM)-1k	& $55.92\%$ \\
OurNetCCE-1k			& $57.44\%$ \\
OurNetCCE(LDA)-1k	& $59.48\%$ \\
DeepLDA-1k			& $\mathbf{66.97\%}$ \\
\end{tabular}
\end{center}
\end{table}

\section{Investigatons on DeepLDA and Discussions}
\label{sec:investigations}
In this section we provide deeper insights into the representations learned by DeepLDA.
We experimentally investigate the eigenvalue structure of representations learned by DeepLDA as well as its relation to the classification potential of the respective networks. 
%!TEX root = iclr2016_conference.tex

\subsection{Does Image Size Affect DeepLDA?}
\begin{figure*}[ht]
\centering
\subfloat[STL-10 ($96\times96$)]{\label{fig:comparison_stl10_32_96}{\includegraphics[width=0.38\textwidth]{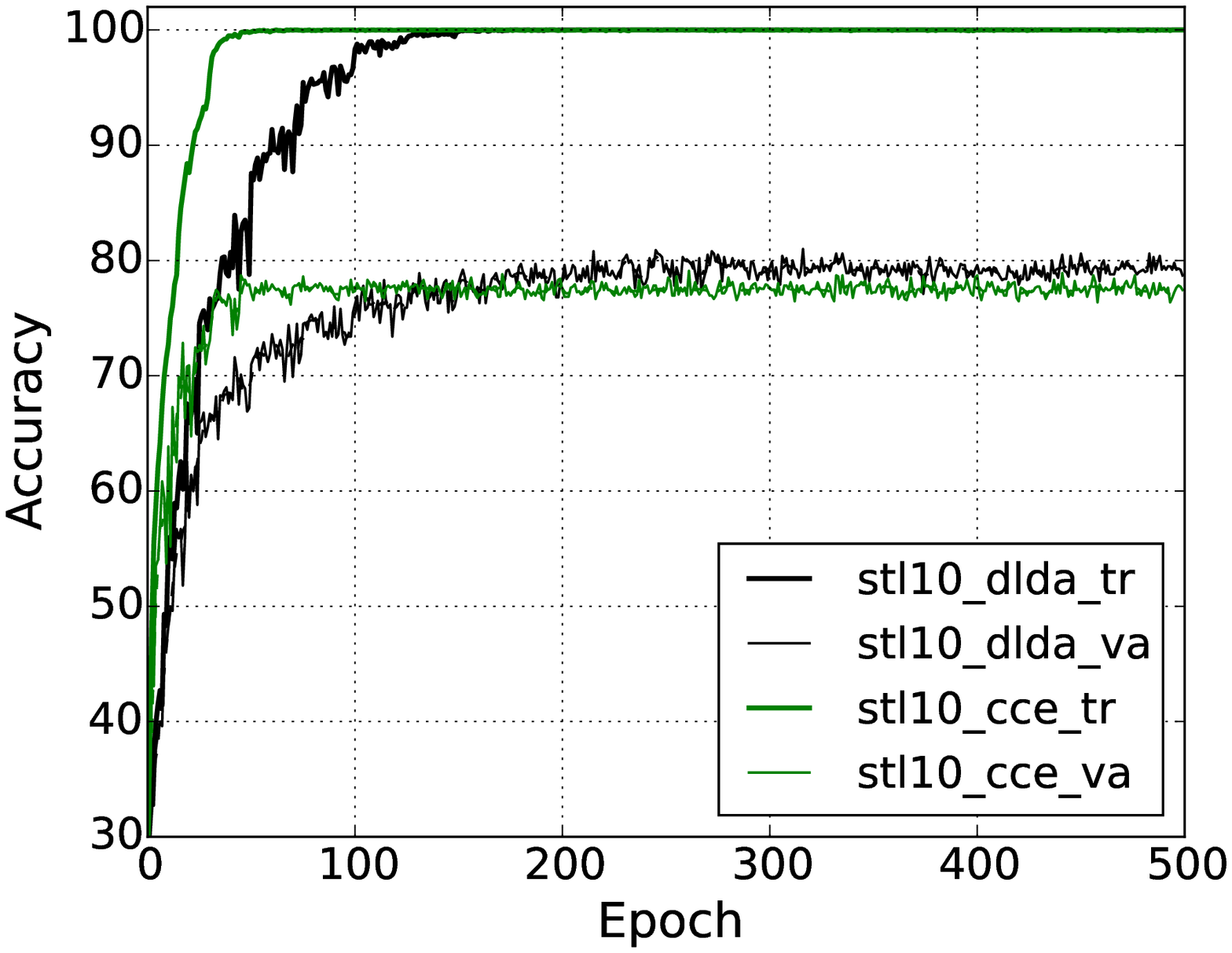} }}%
\qquad
\subfloat[STL-10 ($32\times32$)]{\label{fig:comparison_stl10_32_32}{\includegraphics[width=0.38\textwidth]{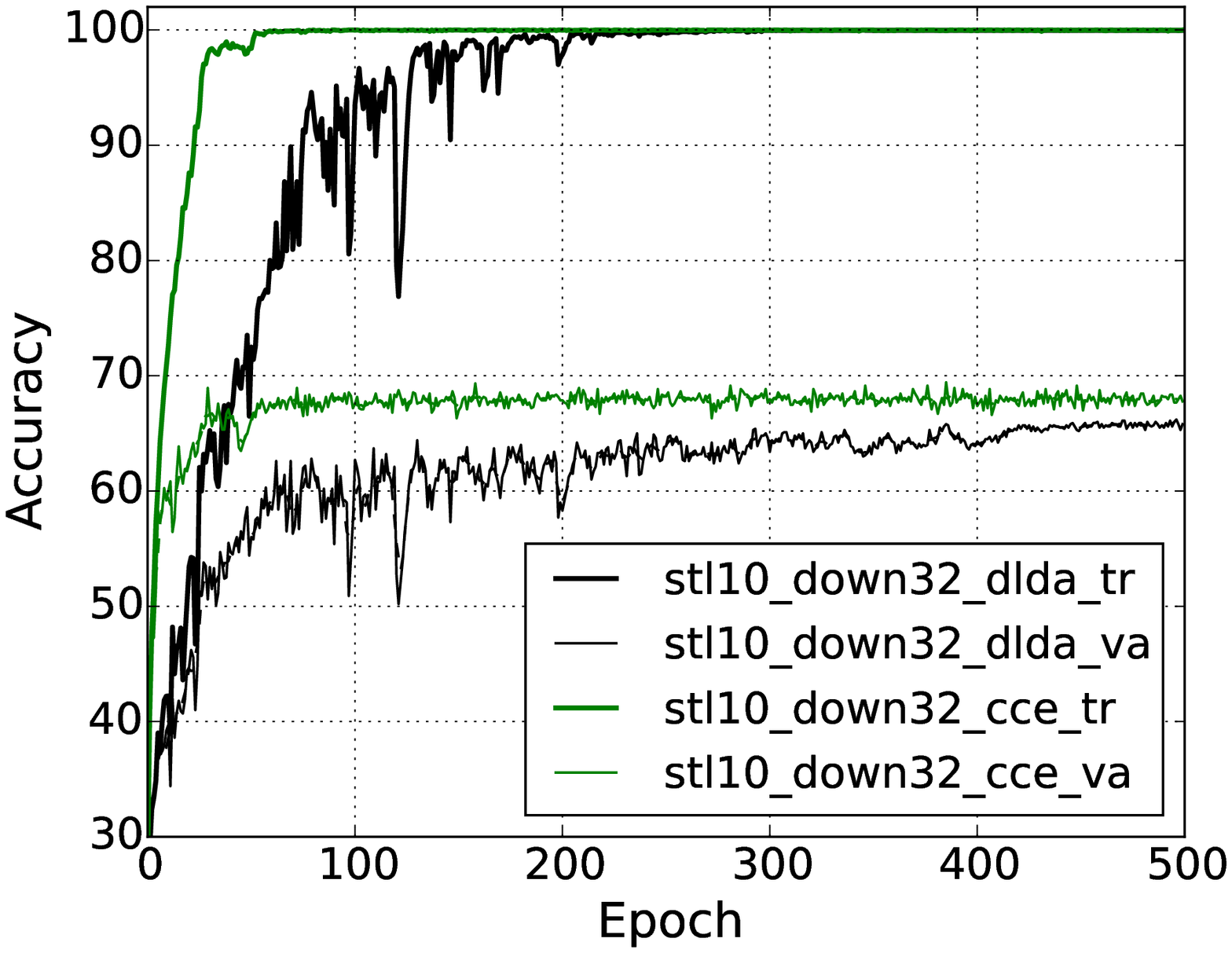} }}%
\caption{Comparison of the learning curves of DeepLDA on the original STL-10 dataset (\emph{Method-4k}) with image size $96\times96$ and its downscaled $32\times32$ version.}
\label{fig:comparison_stl10_32}
\end{figure*}

DeepLDA shows its best performance on the STL-10 dataset (\emph{Method-4k}) where it outperforms CCE by $3$ percentage points.
The major difference between STL-10 and CIFAR-10 -- apart from the number of train images -- is the size of the contained images (see Figure \ref{fig:datasets} to get an impression of the size relations).
To get a deeper insight into the influence of this parameter we run the following additional experiment: (1) we create a downscaled version of the STL-10 dataset with the same image dimensions as CIFAR-10 ($32 \times 32$). (2) We repeat the experiment (\emph{Method-4k}) described in Section \ref{subsec:stl10_results} on the downscaled $32\times32$ dataset.
The results are presented in Figure \ref{fig:comparison_stl10_32}, as curves showing the evolution of train and validation accuracy during training.
As expected, downscaling reduces the performance of both CCE and DeepLDA.
We further observe that DeepLDA performs best when trained on larger images and
has a disadvantage on the small images.
However, a closer look at the results on CIFAR-10 (CCE: $7.10\%$ error, DeepLDA: $7.29\%$ error, see Table \ref{tab:res_cifar10})
suggests that this effect is compensated when the training set size is sufficiently large.
As a reminder: CIFAR-10 contains 50000 train images in contrast to STL-10 with only 4000 samples.

\subsection{Eigenvalue Structure of DeepLDA Representations}
\begin{figure*}[ht]
\centering
\subfloat[Eigenvalues vs. Accuracy]{\label{fig:eigenvalue_structure_accuracy}{\includegraphics[width=0.32\textwidth]{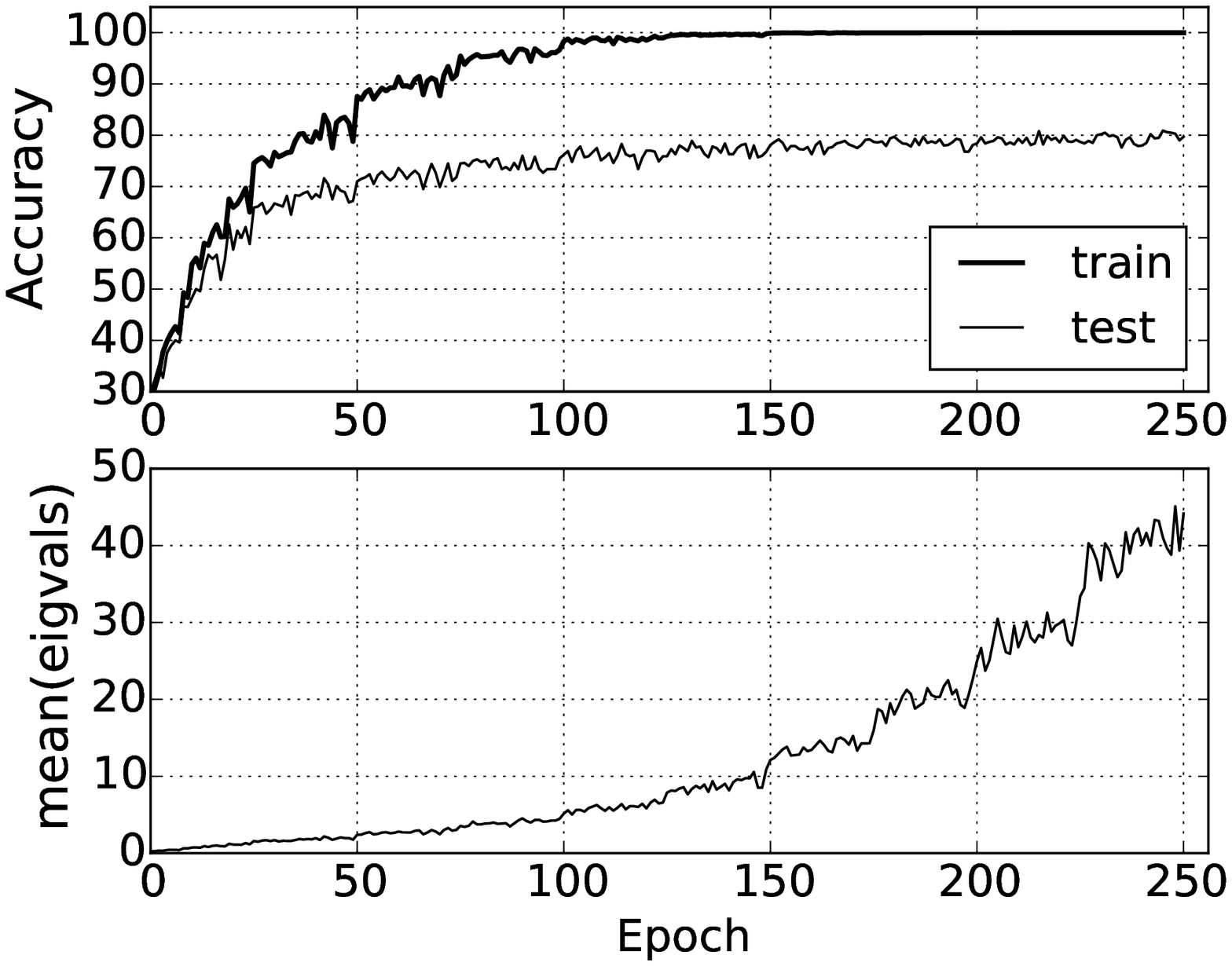} }}%
\subfloat[Individual Eigenvalues]{\label{fig:eigenvalue_structure_evolution}{\includegraphics[width=0.32\textwidth]{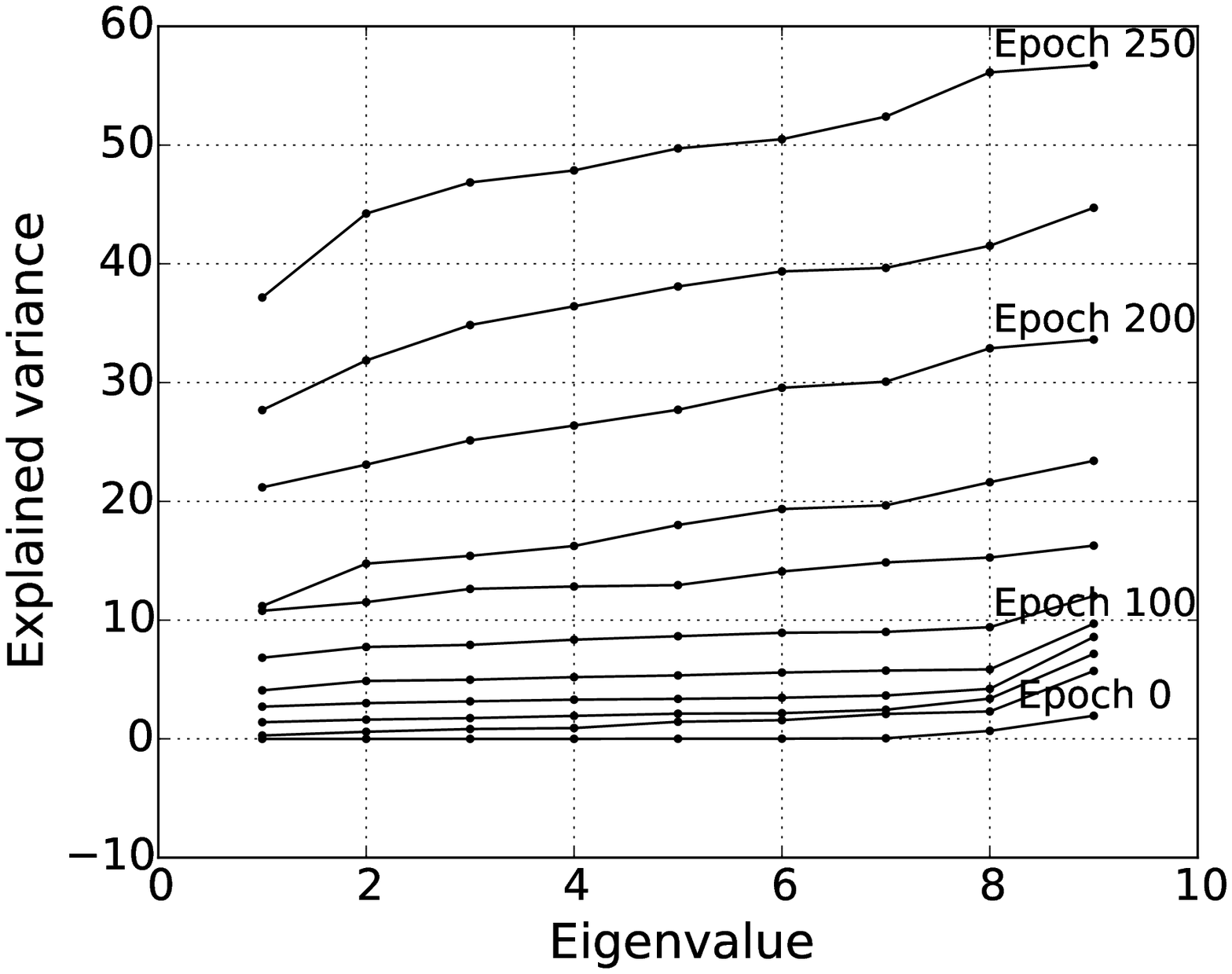} }}%
\subfloat[CCE vs. DeepLDA]{\label{fig:comparison_latent_space}{\includegraphics[width=0.32\textwidth]{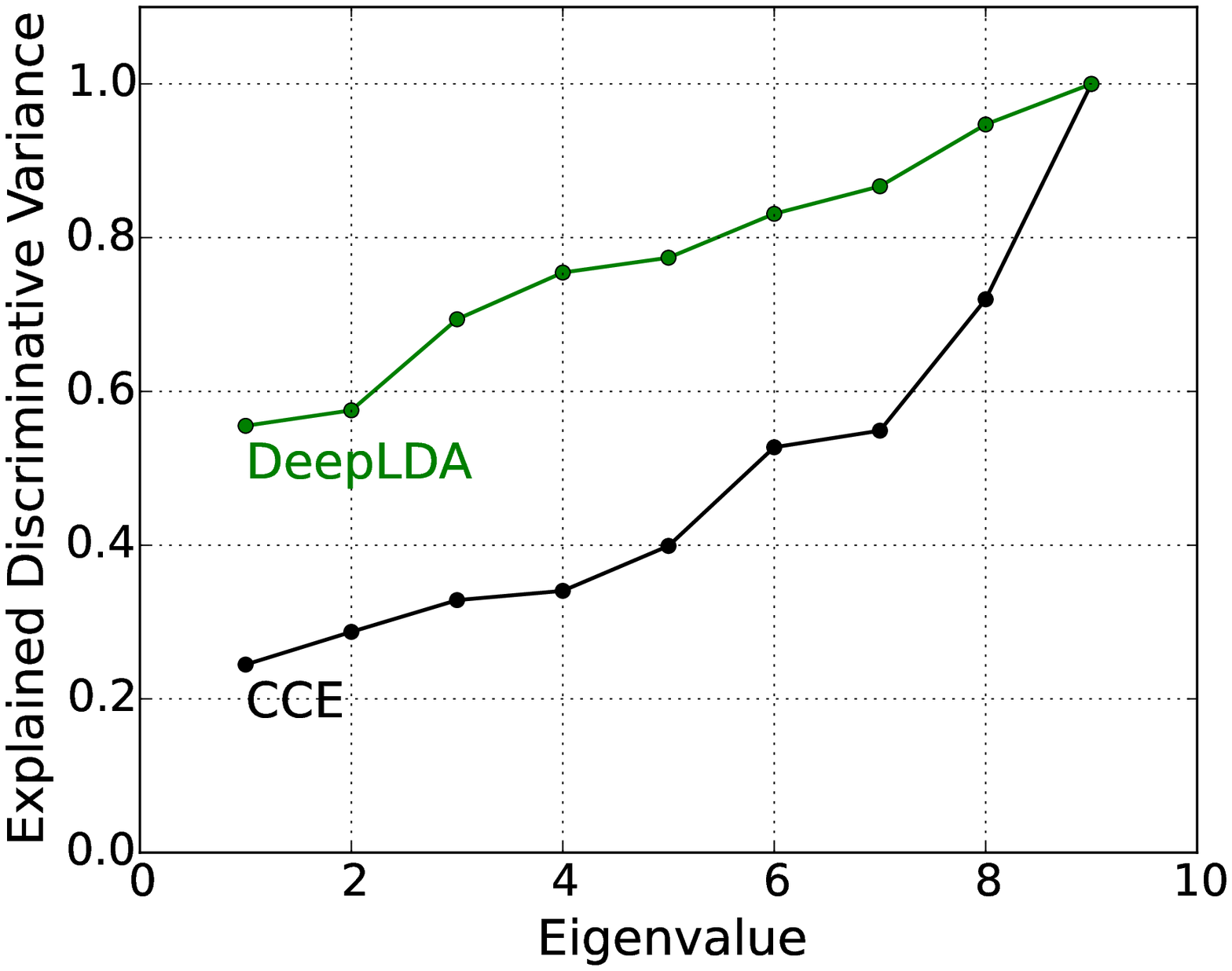} }}%
\caption{The figure investigates the eigenvalue structure of the general LDA eigenvalue problem during training a DeepLDA network on STL-10 (\emph{Method-4k}). (a) shows the evolution of classification accuracy along with the magnitude of explained discriminative variance (separation) in the latent representation of the network.
(b) shows the evolution of individual eigenvalues during training.
In (c) we compare the eigenvalue structure of a net trained with CCE and DeepLDA (for better comparability we normalized the maximum eigenvalue to one).}
\label{fig:eigenvalue_structure}
\end{figure*}

DeepLDA optimization does not focus on maximizing the target class likelihood of individual samples.
As proposed in Section \ref{sec:dlda} we encourage the net to learn feature representations
with discriminative distribution parameters (within and between class scatter).
We achieve this by exploiting the eigenvalue structure of the general LDA eigenvalue problem
and use it as a deep learning objective.
Figure \ref{fig:eigenvalue_structure_accuracy} shows the evolution of train and test set accuracy of STL-10 along with the mean value of all eigenvalues in the respective training epoch.
We observe the expected natural correlation between the magnitude of explained "discriminative" variance (separation) and the classification potential of the resulting representation.
In Figure \ref{fig:eigenvalue_structure_evolution} we show how the individual eigenvalues increase during training. Note that in Epoch 0 almost all eigenvalues (1-7) start at a value of $0$.
This emphasizes the importance of the design of our objective function (compare Equation (\ref{eq:lda_loss})) which allows to draw discriminability into the lower dimensions of the eigen-space.
In Figure \ref{fig:comparison_latent_space} we additionally compare the eigenvalue structure of the latent representation produced by DeepLDA with CCE based training.
Again results show that DeepLDA helps to distribute the discriminative variance
more equally over the available dimensions.
To give the reader an additional intuition on the learned representations we visualize the latent space of STL-10 in our supplemental materials on the final page of this paper.

\section{Conclusion}
\label{sec:conclusion}
%!TEX root = iclr2016_conference.tex

We have presented DeepLDA, a deep neural network interpretation of linear discriminant analysis.
DeepLDA learns linearly separable latent representations in an end-to-end fashion
by maximizing the eigenvalues of the general LDA eigenvalue problem.
Our modified version of the LDA optimization target pushes the network to distribute discriminative variance in all dimensions of the latent feature space.
Experimental results show that representations learned with DeepLDA are discriminative and
have a positive effect on classification accuracy.
Our DeepLDA models achieve competitive results on MNIST and CIFAR-10 and outperform CCE in a fully supervised setting of STL-10 by more than $9\%$ test set accuracy.
The results and further investigations suggest that DeepLDA performs best, when applied to reasonably-sized images (in the present case $96\times96$ pixel).
Finally, we see DeepLDA as a specific instance of a general fruitful strategy:
exploit well-understood machine learning or classification models
such as LDA with certain desirable properties, and use deep networks
to learn representations that provide optimal conditions for these models.

\subsubsection*{Acknowledgments}
We would like to thank Sepp Hochreiter for helpful discussions,
and the three anonymous reviewers for extremely helpful (and partly challenging) remarks.
We would also like to thank all developers of \emph{Theano} \citep{bergstra_2010_theano}
and \emph{Lasagne} \citep{sander_dieleman_2015_27878}
for providing such great deep learning frameworks.
The research reported in this paper has been supported by the Austrian Ministry for Transport, Innovation and Technology, the Federal Ministry of Science, Research and Economy, and the Province of Upper Austria in the frame of the COMET center SCCH.
The Tesla K40 used for this research was donated by the NVIDIA Corporation.

\bibliography{iclr2016_conference}
\bibliographystyle{iclr2016_conference}

\newpage
\section*{Appendix A: Gradient of DeepLDA-Loss}
\label{sec:supplementals_2}
%!TEX root = iclr2016_conference.tex
To train with back-propagation we provide the partial derivatives of optimization target $l(\mathbf{H})$ proposed in Equation (\ref{eq:lda_loss})
with respect to the topmost hidden representation $\mathbf{H}$ (contains samples as rows and features as columns).
As a reminder, the DeepLDA objective focuses on maximizing the $k$ smallest eigenvalues $v_i$ of the generalized LDA eigenvalue problem.
In particular, we consider only the $k$ eigenvalues that do not exceed a certain threshold for optimization:
\begin{equation}
l(\mathbf{H}) = \frac{1}{k} \sum_{i=1}^{k}v_i \; \; \textnormal{with} \; \; \{v_1,...,v_k\} = \{v_j | v_j < \min\{v_1,...,v_{C-1}\} + \epsilon\}
\end{equation}
For convenience, we change the subscripts of the scatter matrices to superscripts in this section (e.g. $\mathbf{S}_t \rightarrow \mathbf{S}^t$).
$\mathbf{S}^t_{ij}$ addresses the element in row $i$ and column $j$ in matrix $\mathbf{S}^t$.
Starting from the formulation of the generalized LDA eigenvalue problem:
\begin{equation}
\mathbf{S}^b \mathbf{e}_i = v_i \mathbf{S}^w \mathbf{e}_i
\end{equation}

the derivative of eigenvalue $v_i$ with respect to hidden representation $\mathbf{H}$ is defined in \citep{Leeuw2007derivatives} as:
\begin{equation}
\frac{\partial v_i}{\partial \mathbf{H}} =
\mathbf{e}_i^T \left( \frac{\partial \mathbf{S}^b}{\partial \mathbf{H}}
- v_i \frac{\partial \mathbf{S}^w}{\partial \mathbf{H}} \right) \mathbf{e}_i
\end{equation}
%given that $\mathbf{S}_w$ is positive definite, which guarantees that all eigenvalues are real, positive and have a multiplicity of $1$. In order for the covariance $\mathbf{S}_w$ to be positive definite, $\mathbf{H}$ may not contain row vectors $\mathbf{h}_i$ that are exact linear combinations of others. This happens to $\mathbf{H}$ with a very low probability, although influenced by mini-batch size. 

Recalling the definitions of the LDA scatter matrices from Section \ref{subsec:lda_general}:
\begin{equation}
\mathbf{S}^c = \frac{1}{N_c-1} \mathbf{\bar{X}}_c^T \mathbf{\bar{X}}_c
\;\;\;\;\;\;\;\;\;\;
\mathbf{S}^w = \frac{1}{C} \sum_{c} \mathbf{S}^c
\end{equation}

\begin{equation}
\mathbf{S}^t = \frac{1}{N-1} \mathbf{\bar{X}}^T \mathbf{\bar{X}}
\;\;\;\;\;\;\;\;\;\;
\mathbf{S}^b=\mathbf{S}^t - \mathbf{S}^w
\end{equation}

we can write the partial derivative of the total scatter matrix $\mathbf{S}_t$ \citep{Andrew2013DCCA, Stuhlsatz2012LDA} on hidden representation $\mathbf{H}$ as:
\begin{equation}
\frac{\partial \mathbf{S}^t_{ab}}{\partial H_{ij}} =
\begin{cases}
    \frac{2}{N-1} \left( H_{ij} - \frac{1}{N} \sum_n H_{nj} \right)	& \text{if } a = j, b = j \\
    \frac{1}{N-1} \left( H_{ib} - \frac{1}{N} \sum_n H_{nb} \right)	& \text{if } a = j, b \neq j \\
    \frac{1}{N-1} \left( H_{ia} - \frac{1}{N} \sum_n H_{na} \right)	& \text{if } a \neq j, b = j \\
    0											              		& \text{if } a \neq j, b \neq j
\end{cases}
\label{eq:deriv_St}
\end{equation}

The derivatives for the individual class covariance matrices $\mathbf{S}^c$ are defined analogously to Equation (\ref{eq:deriv_St}) for the $C$ classes and we can write the partial derivatives of $\mathbf{S}^w$ and $\mathbf{S}^b$ with respect to the latent representation $\mathbf{H}$ as:
\begin{equation}
\frac{\partial \mathbf{S}^w_{ab}}{\partial H_{ij}} = \frac{1}{C} \sum_c \frac{\partial \mathbf{S}^c_{ab}}{\partial H_{ij}}
\;\;\;\;\text{and}\;\;\;\;
\frac{\partial \mathbf{S}^b_{ab}}{\partial H_{ij}} = \frac{\partial \mathbf{S}^t_{ab}}{\partial H_{ij}} - \frac{\partial \mathbf{S}^w_{ab}}{\partial H_{ij}}
\end{equation}

The partial derivative of the loss function introduced in Section \ref{subsec:lda_modified} with respect to hidden state $\mathbf{H}$ is then defined as:
\begin{equation}
\frac{\partial}{\partial \mathbf{H}} \frac{1}{k} \sum_{i=1}^{k}v_i =
\frac{1}{k} \sum_{i=1}^{k} \frac{\partial v_i}{\partial \mathbf{H}} =
\frac{1}{k} \sum_{i=1}^{k} \mathbf{e}_i^T \left( \frac{\partial \mathbf{S}^b}{\partial \mathbf{H}}
- v_i \frac{\partial \mathbf{S}^w}{\partial \mathbf{H}} \right) \mathbf{e}_i
\end{equation}

\newpage
\section*{Appendix B: DeepLDA Latent Representation}
\label{sec:supplementals}
\begin{figure}[h]
\begin{center}
\includegraphics[width=0.95\textwidth]{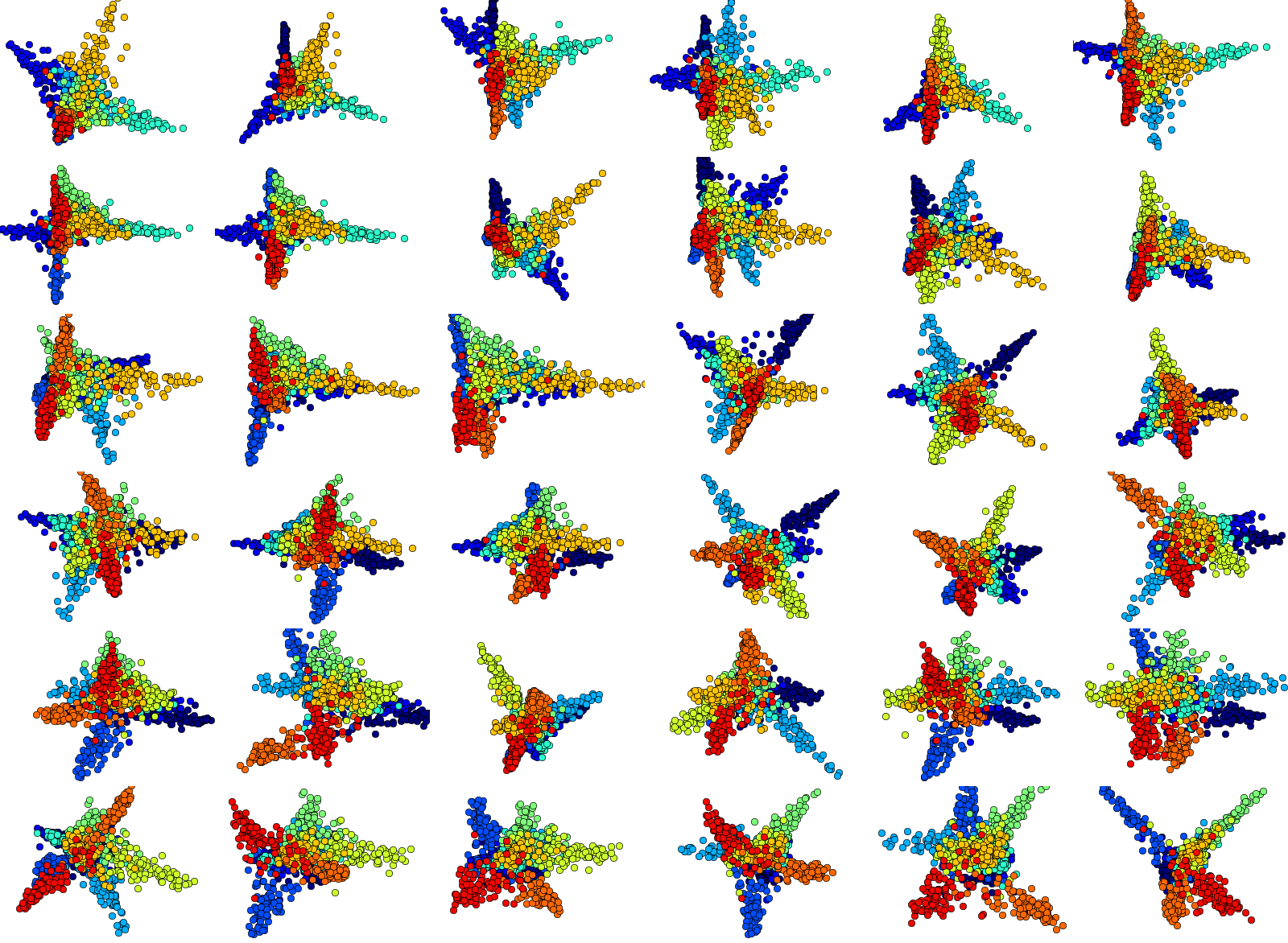}
\end{center}
\caption{STL-10 latent representation produced by DeepLDA (\textit{n-to-n} scatter plots of the latent features of the first 1000 test set samples. e.g.: top left plot: latent feature 1 vs. latent feature 2).}
\label{fig:latent_space_dlda}
\end{figure}

%!TEX root = iclr2016_conference.tex
Figure \ref{fig:latent_space_dlda} shows the latent space representations on the STL-10 dataset (Method-4k)
as \textit{n-to-n} scatter plots of the latent features on the first 1000 test set samples.
We plot the test set samples after projection into the $C-1$ dimensional DeepLDA feature space.
The plot suggest that DeepLDA makes use of all available feature dimensions.
An interesting observation is that many of the internal representations are orthogonal to each other (which is an implication of LDA). This of course favours linear decision boundaries.
%
%For a direct comparison we show in Figure \ref{fig:latent_space_cce} the internal representation of the topmost layer of a network trained with CCE (after Global-Average-Pooling).

%
%\begin{figure}[h]
%\begin{center}
%\includegraphics[width=0.75\textwidth]{figures/stl10_CCE_latent_space.png}
%\end{center}
%\caption{STL-10 latent representation of topmost layer produced by CCE training (\textit{n-to-n} scatter plots of the latent features of the first 1000 test set samples).}
%\label{fig:latent_space_cce}
%\end{figure}

\end{document}